\def\year{2020}\relax
\documentclass[letterpaper]{article} 
\usepackage{aaai20}  
\usepackage{times}  
\usepackage{helvet} 
\usepackage{courier}  
\usepackage[hyphens]{url}  
\usepackage{graphicx} 
\usepackage{graphicx}  
\usepackage{amsmath,amssymb,amsfonts}
\usepackage{algorithmic}
\usepackage{graphicx}
\usepackage{textcomp}
\usepackage{xcolor}
\usepackage[ruled]{algorithm2e}
\usepackage{enumitem}
\setlist[enumerate]{itemsep=0mm}

\usepackage{booktabs} 
\usepackage{multirow}

\title{Self-boosted Time-series Forecasting with Multi-task and Multi-view Learning}

\author{
\Large \textbf{Long H. Nguyen, Zhenhe Pan, Opeyemi Openiyi, Hashim Abu-gellban, Mahdi Moghadasi, Fang Jin}\\ 
\textsuperscript{\rm 1}Department of Computer Science, Texas Tech University\\ 
\{long.nguyen, opeyemi.openiyi, hashim.gellban, mahdi.moghadasi, fang.jin\}@ttu.edu\\
\textsuperscript{\rm 2}Kinetica db Inc.\\ 
zhenhepan@gmail.com
}

\begin{document}

\maketitle

\begin{abstract}
A robust model for time series forecasting is highly important in many domains, including but not limited to financial forecast, air temperature and electricity consumption. To improve forecasting performance, traditional approaches usually require additional feature sets. However, adding more feature sets from different sources of data is not always feasible due to its accessibility limitation. In this paper, we propose a novel self-boosted mechanism in which the original time series is decomposed into multiple time series. These time series played the role of additional features in which the closely related time series group is used to feed into multi-task learning model, and the loosely related group is fed into multi-view learning part to utilize its complementary information. We use three real-world datasets to validate our model and show the superiority of our proposed method over existing state-of-the-art baseline methods.

\end{abstract}

\section{Introduction}
Almost any successful business needs to predict the future in order to make better decisions and allocate resources more effectively. Times series forecasting gives the power of future prediction based on past observations. It is ubiquitous, which is used extensively in finance, supply chain management and inventory planning.  Examples of time series forecasting use cases are: stock market forecasting \cite{ma2018multi},
cotton yield forecasting~\cite{nguyen2019spatial}, gas price forecasting~\cite{jin2015forecasting}, weather forecasting~\cite{wang2019deep},
energy demand forecasting for households ~\cite{chou2018forecasting} and many more.
Time series data is collected at successive and equally-spaced time intervals. Because of the temporal dimension, time series data therefore contains the information from the past to current and possibly conveys the information for future. Accurate time series forecasting algorithms can best capture the observed time series, help interpret the underlying causes, and forecast the future values based on the history of that series.

The undoubted importance of time series forecasting has driven significant interest in this area. \cite{box1970distribution} proposed a statistical method called autoregressive integrated moving average (ARIMA) to forecast univariate time series data. \cite{gers1999learning} proposed Long-Short Term Memory (LSTM) network to learn both long and short term dependency of the data while addressing the gradient vanishing and exploding problem of recurrent neural network (RNN) \cite{hochreiter1998vanishing}. Convolutional neural network in image processing and sequence to sequence in language modeling are also utilized in the task of time series forecast \cite{borovykh2017conditional}, \cite{zaytar2016sequence}. Researchers also endeavor to build more complex models to enhance the forecasting performance \cite{lai2018modeling}. These works have laid the foundations for the time series forecasting, either by utilizing the state-of-the-art deep learning models or heavily relying on domain related knowledge to improve the performance of the forecasting models. 
However, optimizing a complex deep learning network is not an easy task. Besides, domain related knowledge is not always accessible and required a lot of demanding work. Therefore, forecasting time series without domain knowledge and ``auto-optimized" mechanism is still a challenge for researchers. 

On the other hand, multi-task learning \cite{caruana1997multitask} and multi-view learning \cite{sun2013survey} have been introduced with the capability to boost the deep learning model performance. Multi-task learning improves generalization and achieves better efficiency and prediction accuracy by using signals of related tasks as inductive bias. Multi-view learning is able to a set of complement distinct features. Therefore, the views can be employed to comprehensively and accurately describe the data, thus to help improve the learning performance \cite{xu2013survey}.

In this paper, we propose a self-boosted model which combines the learning capability of multi-task and the multi-view learning with co-training objective function to enhance the forecasting performance.
More importantly, the model does not require external knowledge. 
The key idea is the original time series were decomposed into multiple components: the intrinsic mode functions and the residue. 
The decomposition is done via Empirical Ensemble Decomposition (EEMD) algorithm, introduced in signal processing \cite{huang1998empirical}, where the generated signals can have variable amplitude and frequency along the time axis. 
Then, these self-generated time series are clustered into closely related groups and loosely related groups compared with the original time series. The decomposed closely related time series are fed to build a multi-task learning model while the decomposed loosely related time series group is utilized for multi-view learning. This combination helps improve the generalization of the model, and can enhance the performance significantly. Specifically, our contributions in this paper are:

\begin{itemize}
    \item We propose a novel self-boosted mechanism for time series forecasting. It firstly decomposes time series to intrinsic mode functions, then utilizes multi-task and multi-view learning paradigms to build the forecasting model from those generated time series.
    \item To the best of our knowledge, this is one of the first attempts in incorporating EEMD method from signal processing into multi-task and multi-view learning paradigms in addressing time series forecasting problem. 
    \item We demonstrate the superiority of our proposed self-boosted model via extensive experiments on different datasets and compare with the state-of-the-art forecasting techniques. This self-boosted forecasting method is widely applicable to multivariate time series data.
\end{itemize}


\section{Related Work}
\label{sec:related-work}
We review two approaches in addressing time series forecasting problem: the general statistical and machine learning based approaches, and the signal processing based approach.

\subsubsection{Statistical and Machine Learning Based Methods}
Autoregressive Integrated Moving Average (ARIMA) introduced by \cite{box1970distribution} is perhaps a base for many time series forecasting solution. This model is a bundle of two variants: the autoregression(AR) and the moving average (MA) models. One limitation is it only looks back of the dependent variable but fails to capture an unusual change of a pattern.

In another approach Dasgupta et al. used non-linear dynamic boltzmann machines for time series prediction \cite{dasgupta2017nonlinear}. The technique is used to learn a generative model of temporal pattern sequences, using an exact learning rule that maximizes the log likelihood of a given time series. Hsiang-Fu et al. presented temporal graph regularization method in high-dimensional time series forecast \cite{yu2016temporal}. They formed connections to graph regularization methods in the context of learning the dependencies in an autoregressive framework. Support vector machines used in \cite{kim2003financial},\cite{chidlovskii2017multi} for financial time series forecasting and learning of mutually dependent time series in a multi-task setting. Some other works utilized both ARIMA and Multilayer Perceptron (MLP) as in \cite{zhang2003time}, or  \cite{jain2007hybrid} for hydrologic time series forecasting.  \cite{emamgholizadeh2014prediction} used artificial neural network (ANN) with adaptive neuro-fuzzy inference system for ground water level prediction. \cite{liang2018multi} used multi-variable stacked Long-short Term Memory (LSTM) network to learn different time scales and enhance wind speed prediction. However, most of these works focus on high-dimensional time series where domain specific features play an important role in the forecasting models. Instead, we propose a solution based on the novelty of not requiring external information in the forecasting model.

\subsubsection{Signal Processing Based Methods}
The success in signal processing using deep learning captures attention of researchers to apply it for addressing the problem of time series forecasting.  Ena et al. utilized a hybrid model of ARIMA and ANN with Discrete Wavelet Transform (DWT) \cite{khandelwal2015time}. In his work DWT is used to decompose a time series dataset into linear and nonlinear components; in the later phase the ARIMA and ANN were used to perform better prediction on those linear and non-linear components, respectively. \cite{awajan2018improving} uses EMD-HW (Ensemble Mode Decomposition - Holt-Winter) bagging technique to do financial forecasting. \cite{wu2019improved} combined ensemble empirical mode decomposition (EEMD) with the LSTM model to forecast crude oil price. These models have shown signal processing success in time series forecasting. However, there are still limited works in integrating the success of latest signal processing technologies and the multi-task multi-view learning paradigms. Our paper also proposes the use of EEMD from signal processing in multi-task multi-view deep neural network setting to bridge the gap.

\section{Problem Formulation}
\label{sec:problem_formulation}
Here, we describe our self-boosted time series forecasting problem as below:

\textbf{Intrinsic mode functions (IMF)}: 
\textit{``IMF is any time-varying function with the same number of extrema and zero crossings, whose envelopes are symmetric with respect to zero"} \cite{huang1998empirical}. 

\textbf{The input}: The input is a univariate time series denoted by $Y=\{y_1, y_2,...,y_t\}$ where $t$ is the current time. 
This time series is decomposed into intrinsic mode functions denoted $IMF=\{IMF_1, IMF_2,..., IMF_n\}$ where $n$ is the number of intrinsic mode functions. Each function $IMF_i$ is described as a time series $IMF_i=\{imf_1, imf_2, ..., imf_t\}$. We call these functions as supporting time series.

\textbf{Problem definition:} With the given time series and its intrinsic mode functions, our goal is to learn a function $f$ that takes the history of those functions and the time series until current time $t$ then returns the predicted values in future time steps $Y_{t+1..t+H} = \{y_{t+1}, y_{t+2}, ..., y_{t+H}\}$ where $H$ is called the forecasting horizon.
\begin{equation}
    y_{t+1}, y_{t+2}, ..., y_{t+H} = f(IMF_{1..n}, Y_{1..t})
\end{equation}

\section{Method}
\label{sec:method}
Here we describe the details of our proposed method. The overall the framework involves three steps:
\begin{itemize}
   \item \textbf{Time series Decomposition:} In this step, the original time series is decomposed into multiple intrinsic mode functions $\{IMF_1, IMF_2,...,IMF_n\}$ in which they are orthogonal and its sum is the original time series. 
   \item \textbf{Feature Selection:} We treat each intrinsic mode function as an additional feature to train our model later. We group all intrinsic mode functions that are similar most with the original time series and build multi-task model based on these features. The similarity measure is calculated via the correlation coefficient between two time series.
   \item \textbf{Forecasting Model:} Finally, we build multi-task model where each task is the forecasting of the selected intrinsic mode functions. The rest of intrinsic mode functions play as additional views for the target task specific branch. The utilization of the co-training algorithm in multi-task learning and the multi-view learning help boost the performance of the forecasting model.
\end{itemize}



\begin{figure}[t]
    \centering
    \includegraphics[width=\linewidth]{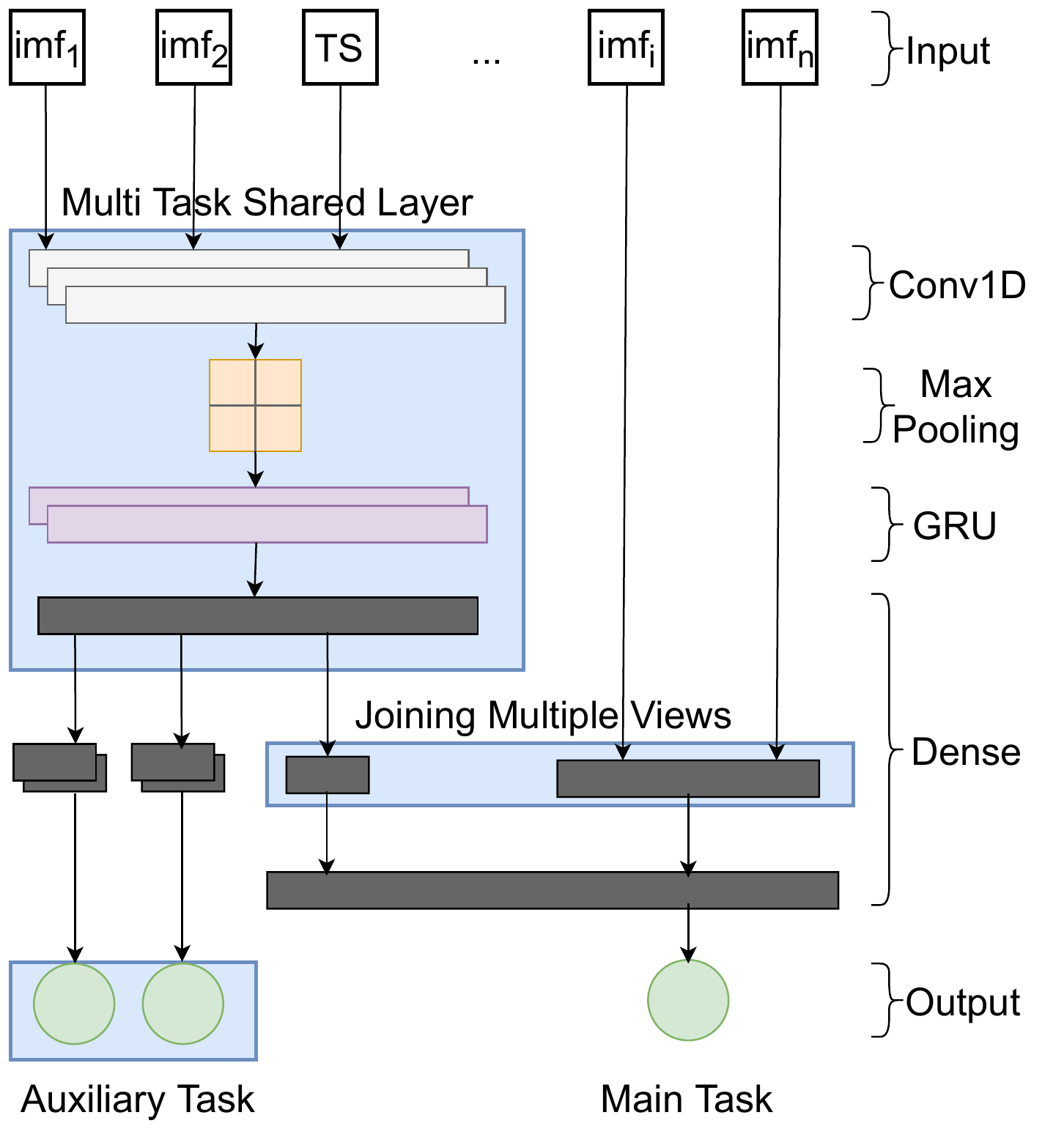}
    \caption{An overview of the multi-task and multi-view learning architecture in self-boosted time series forecasting framework.}
    \label{fig:architecture}
\end{figure}




\subsection{From Time series to Intrinsic Mode Functions}
We employ Ensemble Empirical Mode Decomposition (EEMD) method from signal processing domain to decompose the input sequence $y(t)$ into serial components (so called intrinsic mode functions (IMF) and residual component) as shown in equation (\ref{equation:ts-decomposition}). 

\begin{equation}
\label{equation:ts-decomposition}
    y(t) = \sum_{i=1}^{N-1}{imf_i} + r_n
\end{equation}
where $imf_i$ is an IMF and $r_n$ is the residual component.
All the IMFs are orthogonal and their sum is equal to the original time series. Each IMF represents a unique range of energy and frequency. 

\begin{algorithm}

    \SetKwInOut{Input}{Input}
    \SetKwInOut{Output}{Output}

    \underline{function eemd\_decomposition} $(y)$\\
    \Input{time series $y$}
    \Output{serial components $imfs$}
     - 0. add uniform distribution of white noise signal. \\
     - 1. identify all extrema of $Y$\\
     - 2. interpolate between minima (resp. maxima) with "envelops" \\
     - 3. compute the mean envelops\\
     - 4. extract the detail $imf_j=y(t)-m_k(t)$ \\
     - 5. repeat (1) to (4) until IMFs meet the definition and converge.\\
     - 6. repeat (1) to (5) to generate residual $r_n(t)=x(t)-imf_n(t)$
     
     \Return $imf_{1..n}$
    
\caption{EEMD decomposition algorithm. Step $1$ to $6$ is the algorithm of Empirical Mode Decomposition (EMD). Step $0$ is added to resolve the mode mixing problem.}
\label{algorithm:eemd-decomposition}
\end{algorithm}

\paragraph{EEMD algorithm.}
Let $N$ be the number of ensembles. The final IMFs and residual in Algorithm \ref{algorithm:eemd-decomposition} are defined by the average values after $N$ ensembles as the following equations:
\begin{equation}
\begin{aligned}
& imf_j(t) = \frac{1}{N} \sum_{i=1}^{N}{imf_{ji}} \quad  j=1..N-1\\
& r(t) = \frac{1}{N} \sum_{i=1}^{N}{r_{ni}}
\end{aligned}
\end{equation}

The stopping condition for Algorithm \ref{algorithm:eemd-decomposition} is the size of the standard deviation (SD) by twice sifting the results \cite{huang1998empirical}. The $SD$ is computed as:
\begin{equation}
    SD = \sum_{t=0}^{T}{ \frac{|imf_{j-1}(t) - imf_j(t)|^2}{imf^2_{j-1}(t)} }
\end{equation}
According to \cite{huang1998empirical}, a typical value of $SD$ is between $0.2$ and $0.3$. The sifting process will be terminated once the $SD$ value falls into the specified range. 

\subsection{Selection of Intrinsic Mode Functions}
To integrate intrinsic mode functions into our multi-task learning model, we perform clustering of those functions into two categories. The first category comprises of functions that are not highly dependent with the original time series, the second category includes functions which are highly dependent with the original one. We employ k-mean algorithm to cluster the time series where $k=M$ ($M$ is the number of clusters) and the distance is measured by the inversion of the similarity (correlation coefficient) between a series and the original one. The general equation for measuring the correlation coefficient between time series $X$ and $Y$ is:
\begin{equation}
    corr(X, Y) = \frac{ \sum_{i=1}^{T}{(x_i-\overline{x})*(y_i-\overline{y})}}{\sqrt{\sum_{i=1}^{T}{(x_i-\overline{x})}}*\sqrt{\sum_{i=1}^{T}{(y_i - \overline{y})}}}
\end{equation}
where $\overline{x}$ and $\overline{y}$ is the mean value of the time series $X$ and $Y$ respectively. Each cluster represents how close it is with the original time series. The least similarity group(s) can be dropped to reduce the data dimension and the computation of the model.

\subsection{Utilizing Intrinsic Mode Functions to Learn with Multi-task and Multi-view based Model}

Figure \ref{fig:architecture} presents an overview of our proposed architecture. The main task in our model is to forecast the original time series at a horizon $H$. The auxiliary tasks are to forecast related time series (intrinsic mode functions) decomposed in the previous step. The main task sub-network takes less related time series group as different views to enhance its forecasting performance. These tasks are co-trained in order to boost the performance of the main task leveraging the auxiliary tasks.

In particular, the related time series group is fed from the input to three one-dimension convolutional layers to extract short-term patterns of the series. Let $k$ be the number of filters with width $w$ swept through the input matrix $X$, the output $h_k$ of each layer after the filter $k^{th}$ is computed as:
\begin{equation}
    h_k = max(0, W_k*X + b_k)
\end{equation}
where $b_k$ is the bias in the filter $k^{th}$ operation; $max(0, x)$ is the popular $RELU(x)$ activation function. Stacking the convolutional layers allows a hierarchical decomposition of the inputs.

A max pooling layer is stacked on top of the convolutional layers to reduce the latent representation dimension and computation in the network. Subsequently, the architecture continues with two gated recurrent units (GRU) \cite{chung2014empirical} and a dense layer before leading into task specific branches. The choice of GRU is for faster training while still maintaining the capabilities of learning long-short term pattern as in the traditional Long-short Term Memory (LSTM) unit. Stacking the GRUs helps efficiently discover more high-level features at different time scales and result in improvement of the forecasting performance.   
The hidden state of recurrent units is computed at time $t$ as below \cite{chung2014empirical}:
\begin{equation}
\begin{aligned}
& h_t = (1-z_t)h_{t-1} + z_t \hat{h_t} \\
& z_t = \sigma(W_z x_t + U_z h_{t-1}) \\
& \hat{h_t} = tanh(W x_t + U(r_t \odot h_{t-1})) \\
& r_t = \sigma(W_r x_t + U_r h_{t-1})
\end{aligned}
\end{equation}
where $r$, $z$, $h$ and $\hat{h}$ are the values of the reset gate, update gate, activation gate and candidate activation, respectively. $\odot$ is the element-wise product, $\sigma$ is the sigmoid function and $x_t$ is the input at time $t$. $W$ and $U$ are weight matrices.

On the main task specific branch, the less related time series group is treated as different views and they are concatenated together before feeding to the subsequent layers. The concatenated view is defined by:
\begin{equation}
    h^c = h^m \oplus v_1 \oplus \dots \oplus v_q
\end{equation}
where $\oplus$ is the concatenation operator, $h^c$ is the final concatenated view, $h^m$ is the latent view of the main task in multi-task learning, $v1 \dots v_q$ are additional views fed by less related time series group.

All the branches end with a fully connected dense layer where it produces the final forecasting results $o_t$ computed as:
\begin{equation}
    o_t = W h_{t} + b
\end{equation}
where $W$ and $b$ are learnable parameters, $h_t$ is the hidden state at time $t$.

\subsection{Optimization Algorithm}
We use traditional strategy in building features and solving time series model. From the given time series $Y_t = \{y_1, y_2, \dots, y_t\}$ and a lag time $q$, we form the input with features are the values $X = \{y_{t- q+1}, y_{t-q+2}, \dots, y_t\}$. If $H$ is the forecasting horizon, the regression problem of feature-value pair $\{X_t, Y_{t+H}\}$ can be solved using Adam optimizer \cite{kingma2014adam}.

Regarding the objective function, it is to minimize the joint loss of all tasks. With this joint loss function, all the tasks are co-trained to improve its generalization. In particular, the joint loss function $L$ is defined by the average weighted loss of all task-specific losses. 
\begin{equation}
    L = \frac{1}{N} \sum_{i=1}^{N} \alpha_i * MSE(Y_i, \hat{Y_i})
\end{equation}
where $N$ is the number of tasks, $\alpha_i$ is the loss weight of the task $i$; $Y_i$ and $\hat{Y_i}$ is the ground truth and forecasting values of all samples in the training set for the task $i$. In our experiment, we penalize more with the error on the main task, hence we set its weight twice compared with the auxiliary tasks.

\begin{figure}[t]
    \centering
    \includegraphics[width=\linewidth]{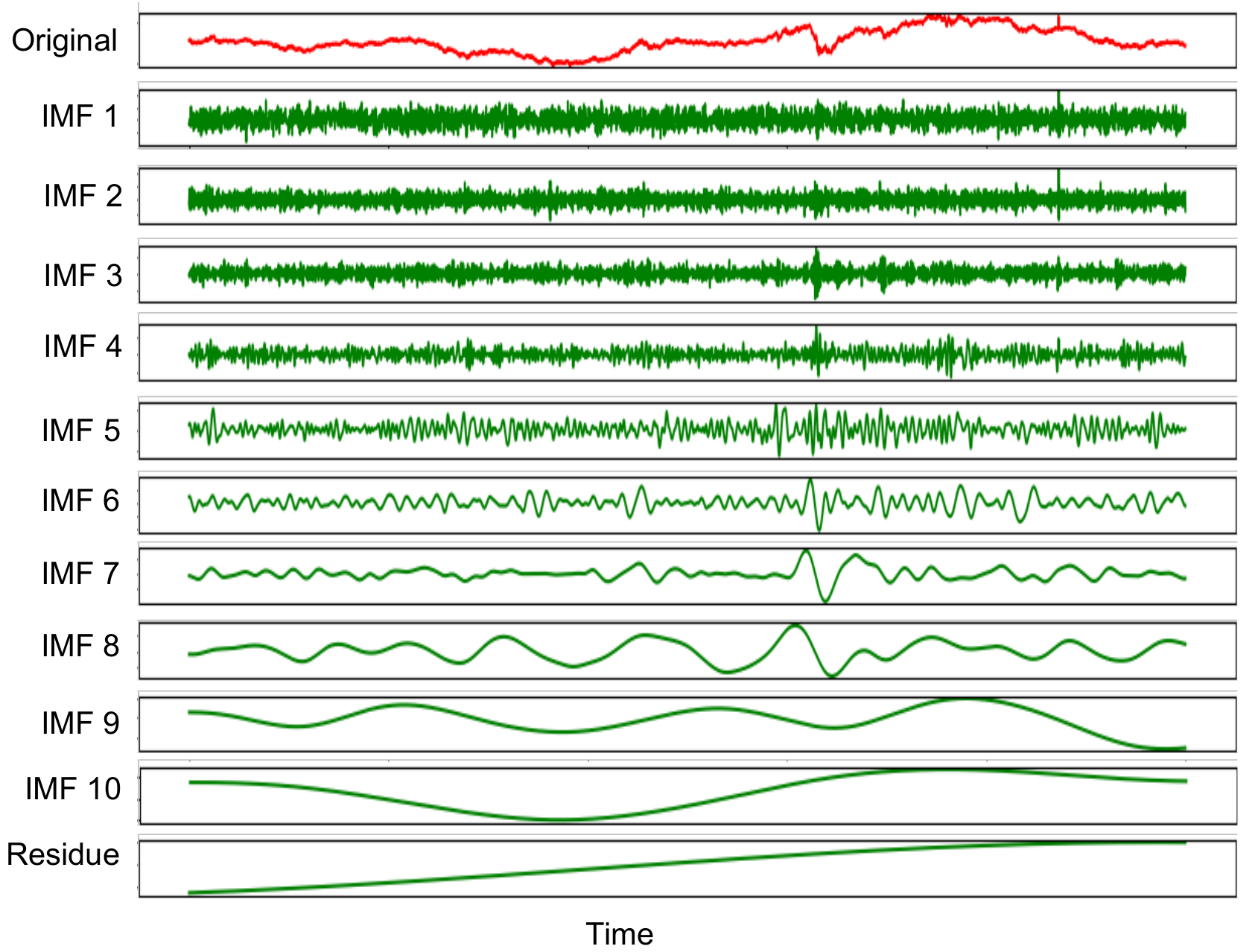}
    \caption{Exchange rate decomposed into intrinsic mode functions.}
    \label{fig:imfs}
\end{figure}

\begin{figure*}[t]
	\centering
	 \includegraphics[width=\linewidth]{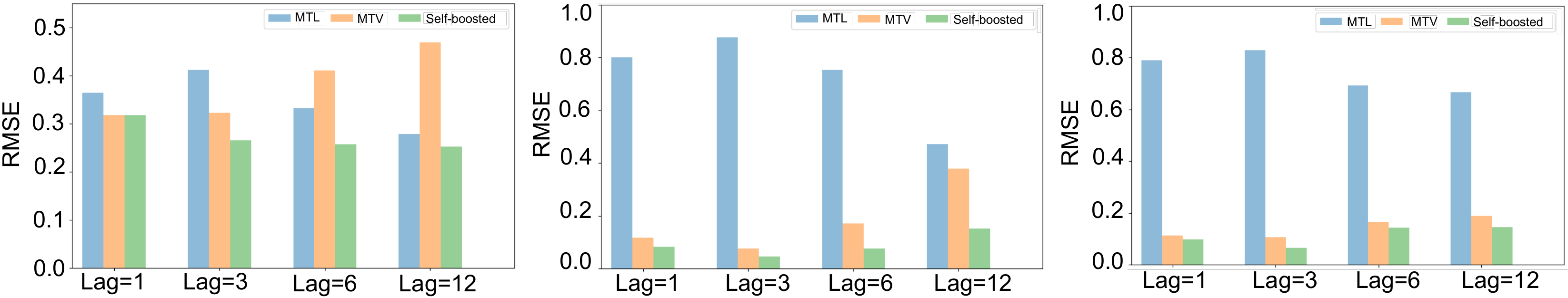}
	\caption{Performance comparison of multi-task learning (MTL), multi-view Learning (MTV) and the proposed method across all datasets (left: electricity consumption; middle: air temperature; right: exchange rate). X-axis is the lag time and Y-axis is the normalized RMSE value.}
\label{fig:comparison-mtl-mtv-all}
\end{figure*}

\section{Evaluation}
\label{sec:evaluation}
We conduct experiments on three public datasets with four state-of-the-art methods. The competing approaches and evaluation metrics are described as below:
\subsection{Competing Approaches}
\begin{itemize}
    \item \textit{ARIMA}: The autoregressive integrated moving average (ARIMA) is a popular time series analysis method, applying in many application domains, including but not limited to primary energy demand \cite{ediger2007arima}. This forecasting technique projects the future values of a series based entirely on its own inertia.
    \item \textit{RNN-GRU}: This is the Recurrent Neural Network (RNN) using Gate Recurrent Unit (GRU) \cite{chung2014empirical} as the cell. The GRU has less number of parameters compared to LSTM while still maintaining competitive performance.
    \item \textit{Dilated CNN}: This is a Convolution Neural Network (CNN) based on WaveNet architecture \cite{oord2016wavenet}. It comprises three dilated convolutional layers. The dilation values for the layers are $1$, $2$, and $4$ respectively \cite{borovykh2017conditional}. 
    \item \textit{Seq2seq}: This is the sequence to sequence method which is widely used in neural machine translation \cite{sutskever2014sequence}. We utilize this approach in time series forecasting in which two Long-Short Term Memory networks (LSTM) are used. One plays the role of encoder while the other one is the decoder.
\end{itemize}

\subsection{Evaluation Metrics}
We use conventional evaluation metrics such as Root Mean Square Error (RMSE), Mean Absolute Error (MAE), Mean Absolute Percentage Error (MAPE) and R-squared ($R^2$). These metrics are computed as: 
\begin{equation}
\begin{aligned}
& RMSE = \sqrt{ \frac{1}{N} \sum_{i=1}^{N} (Y_i - \hat{Y_i})^2} \\
& MAE = \frac{1}{N} \sum_{i=1}^{N} |Y_i - \hat{Y_i}| \\
& MAPE = \frac{100\%}{N} \sum_{i=1}^{N} \frac{|Y_i - \hat{Y_i}|}{Y_i}
\end{aligned}
\end{equation}
where $i = 1,\dots,N$; $Y$ and $\hat{Y}$ are ground true series and system forecast series. $N$ is the number of elements in the test set.

\section{Experiment Results}
\label{sec:results}

\begin{table*}[t]
  \centering
\caption{Summary of all models' performance. Each row is the metric values of a specific method for all datasets. Each column is the comparison between methods on a particular dataset using metrics RMSE, MAE, MAPE and R-squared (R2). Report for lag time $1$ is not included due to limited space.}     \begin{tabular}{rlrrrrrrrrr}
    \toprule
    \multicolumn{1}{l}{Dataset} & \multicolumn{1}{r|}{} & \multicolumn{3}{c|}{Electricity Consumption} & \multicolumn{3}{c|}{Air Temperature} & \multicolumn{3}{c}{Exchange Rates} \\
    \midrule
          & \multicolumn{1}{r|}{} & \multicolumn{3}{c|}{Lag Time} & \multicolumn{3}{c|}{Lag Time} & \multicolumn{3}{c}{Lag Time} \\
\cmidrule{3-11}    \multicolumn{1}{l}{Method} & \multicolumn{1}{l|}{Metrics} & \multicolumn{1}{c}{3} & \multicolumn{1}{c}{6} & \multicolumn{1}{c|}{12} & \multicolumn{1}{c}{3} & \multicolumn{1}{c}{6} & \multicolumn{1}{c|}{12} & \multicolumn{1}{c}{3} & \multicolumn{1}{c}{6} & \multicolumn{1}{c}{12} \\
    \midrule
    \multicolumn{1}{r}{\multirow{4}[2]{*}{ARIMA}} & RMSE  & 268.2087 & 263.7740 & 256.1630 & 7.0773 & 7.0090 & 6.8422 & 0.0130 & 0.0130 & 0.0130 \\
          & MAE   & 196.6700 & 193.1916 & 186.7810 & 4.7797 & 4.7365 & 4.6165 & 0.0094 & 0.0094 & 0.0094 \\
          & MAPE  & 0.3820 & 0.3763 & 0.3646 & 0.0922 & 0.0914 & 0.0892 & 0.0104 & 0.0104 & 0.0104 \\
          & R2    & -0.4037 & -0.3577 & -0.2804 & 0.4079 & 0.4193 & 0.4466 & 0.9871 & 0.9871 & 0.9871 \\
    \midrule
    \multicolumn{1}{r}{\multirow{4}[2]{*}{RNN-GRU}} & RMSE  & 164.1796 & 272.0544 & 188.6766 & 1.8021 & 1.7469 & \textbf{1.8587} & 0.0072 & 0.0106 & 0.0114 \\
          & MAE   & 115.5435 & 189.5071 & 159.8048 & 1.2333 & 1.2486 & \textbf{1.3186} & 0.0041 & 0.0071 & 0.0068 \\
          & MAPE  & 0.0657 & 0.0980 & 0.0875 & 0.0248 & 0.0254 & \textbf{0.0265} & 0.0048 & 0.0079 & 0.0073 \\
          & R2    & 0.9684 & 0.9131 & 0.9582 & 0.9713 & 0.9731 & \textbf{0.9696} & 0.9957 & 0.9906 & 0.9890 \\
    \midrule
    \multicolumn{1}{r}{\multirow{4}[2]{*}{Dilated CNN}} & RMSE  & 152.0230 & 123.9416 & 111.9258 & 3.2532 & 6.5757 & 4.3498 & \textbf{0.0064} & 0.0134 & 0.0121 \\
          & MAE   & 87.2483 & 30.0819 & 79.2819 & 2.5077 & 4.3453 & 3.1238 & \textbf{0.0038} & 0.0075 & 0.0086 \\
          & MAPE  & 0.0505 & 0.0629 & 0.0490 & 0.0583 & 0.1068 & 0.0734 & \textbf{0.0044} & 0.0081 & 0.0097 \\
          & R2    & 0.9820 & 0.9660 & 0.9853 & 0.9066 & 0.6189 & 0.8336 & \textbf{0.9966} & 0.9850 & 0.9878 \\
    \midrule
    \multicolumn{1}{r}{\multirow{4}[2]{*}{Seq2seq}} & RMSE  & 151.6445 & 142.9175 & 129.0600 & 1.7772 & 2.0277 & 2.1882 & 0.0144 & 0.0168 & 0.0146 \\
          & MAE   & 104.5271 & 99.3870 & 97.1598 & 1.2675 & 1.6001 & 1.7653 & 0.0100 & 0.0133 & 0.0094 \\
          & MAPE  & 0.0571 & 0.0551 & 0.0522 & 0.0258 & 0.0336 & 0.0373 & 0.0108 & 0.0148 & 0.0101 \\
          & R2    & 0.9730 & 0.9760 & 0.9804 & 0.9721 & 0.9638 & 0.9579 & 0.9827 & 0.9763 & 0.9820 \\
    \midrule
    \multicolumn{1}{r}{\multirow{4}[2]{*}{\textbf{Self-boosted}}} & RMSE  & \textbf{98.4757} & \textbf{85.9606} & \textbf{88.9537} & \textbf{1.4645} & \textbf{1.3909} & 1.9176 & 0.0070 & \textbf{0.0096} & \textbf{0.0082} \\
          & MAE   & \textbf{73.4041} & \textbf{62.3544} & \textbf{65.4086} & \textbf{1.1418} & \textbf{1.0817} & 1.5684 & 0.0046 & \textbf{0.0071} & \textbf{0.0058} \\
          & MAPE  & \textbf{0.0405} & \textbf{0.0363} & \textbf{0.0398} & \textbf{0.0234} & \textbf{0.0221} & 0.0324 & 0.0054 & \textbf{0.0088} & \textbf{0.0067} \\
          & R2    & \textbf{0.9886} & \textbf{0.9913} & \textbf{0.9907} & \textbf{0.9811} & \textbf{0.9830} & 0.9677 & 0.9960 & \textbf{0.9924} & \textbf{0.9944} \\
    \bottomrule
    \end{tabular}%
  \label{tab:performance-comparison}%
\end{table*}%

\subsection{Dataset Description}
We use publicly available datasets which can be summarize as below:

\begin{itemize}
   \item \textit{Electricity\footnote{https://archive.ics.uci.edu/ml/datasets/ElectricityLoadDiagrams20112014}}: This is the electricity consumption in kW was recorded
every $15$ minutes from $2011$ to $2014$, for $321$ clients. We resampled for  hourly consumption and average them to have per client consumption. This data is used for training model that forecasts average electricity per client consumption.
   \item \textit{Exchange rate}: The dataset contains daily exchange rates from $1990$ to $2016$ of eight countries: Australia, British, Canada, Switzerland, China, Japan, New Zealand and Singapore.
   
   \item \textit{Air Temperature\footnote{https://archive.ics.uci.edu/ml/datasets/Air+Quality}}: This is the hourly temperature recorded from March 2004 to February 2005 on the field within an Italian city. The missing values are replaced with linear interpolation.
\end{itemize}
All the datasets are split into $60\%$, $20\%$, and $20\%$ within chronological order for training, validation and testing, respectively. All the models forecast one-time step forward.

\subsection{Ensemble Empirical Mode Decomposition Result}
Figure \ref{fig:imfs} presents the IMFs after decomposing the original exchange rate time series using EEMD method. There are $10$ IMF components and one residue. The $IMF1$ has the highest frequency, shortest wavelength and maximum amplitude. The subsequent components have the decreasing frequency, amplitude and increasing wavelength. The residual component has a slowly varying around long term representing the trend of annual exchange rate pattern. The EEMD decomposition transforms non-linear, non-stationary time series to stationary time series and can be useful for forecasting performance.

\begin{figure}[t]
    \centering
    \includegraphics[width=\linewidth]{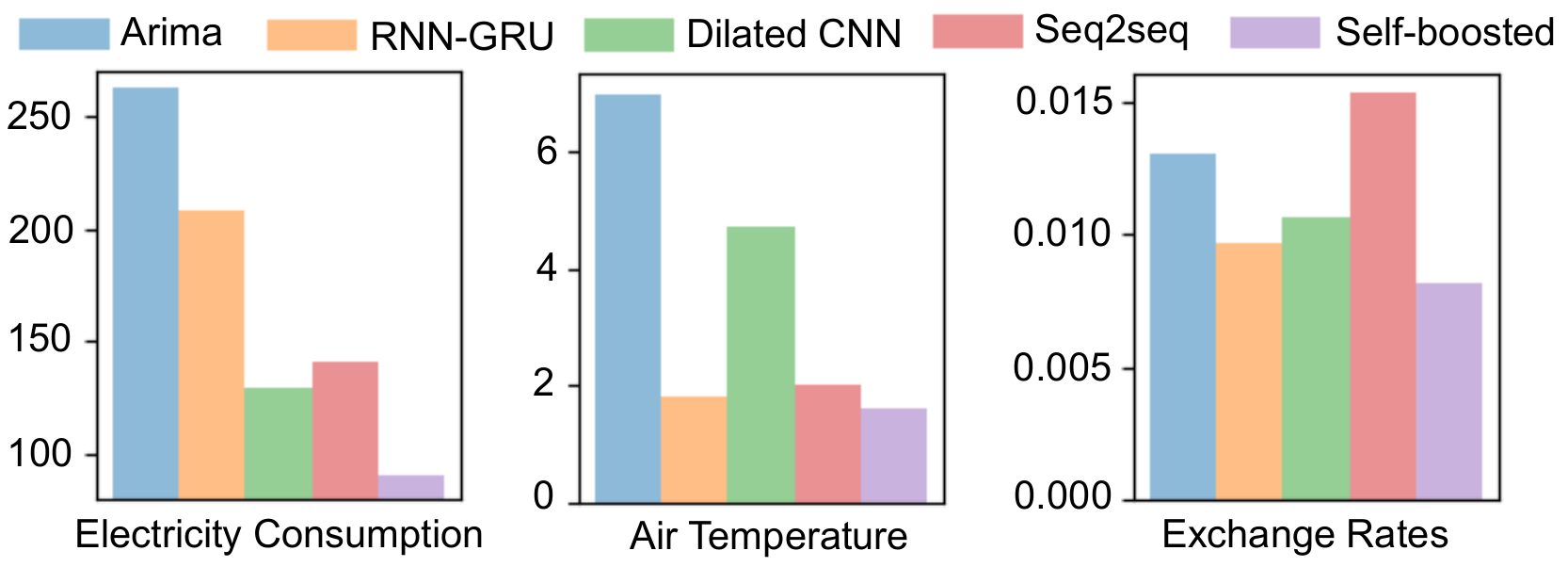}
    \caption{Average RMSE comparison on each dataset.
    }
    \label{fig:avg-rmse-comparison}
\end{figure}

\begin{figure*}[t]
    \centering
    \includegraphics[width=\linewidth]{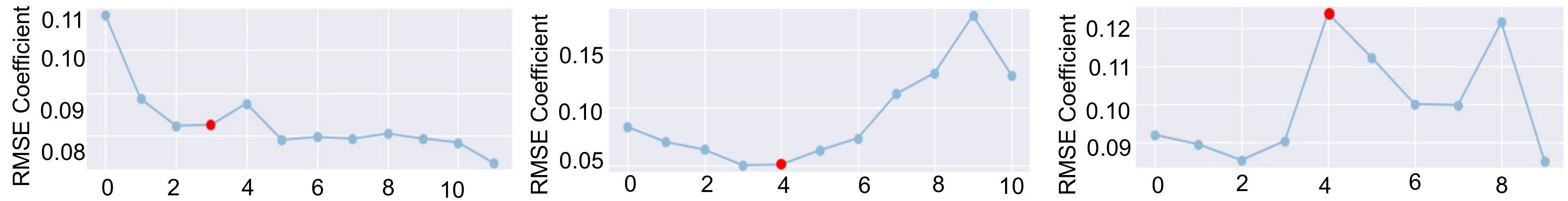}
    \caption{Impact of intrinsic mode functions on model performance across three datasets (left: electricity consumption, middle: air temperature, right: exchange rates). X-axis indicates the number of intrinsic mode functions. Y-axis is the RMSE coefficient. The red point indicates starting of adding intrinsic mode functions to the multi-view learning component.
    }
    \label{fig:performance-imfs}
\end{figure*}

\subsection{Overall Forecasting Performance}
Table \ref{tab:performance-comparison} summarizes the forecasting performance of our proposed method comparing with all the baselines across three datasets and all the metrics. We also vary the lag time with values $\{1, 3, 6, 12\}$ to evaluate the learning capabilities of the models better. Due to limited space, the table does not show the result of the lag time $1$. The best results are highlighted in bold face.

From Table \ref{tab:performance-comparison}, we see that deep learning models outperforms the traditional ARIMA model in all perspectives. This explains the skill of deep learning models in learning both linear and non-linear time series data. For electricity consumption dataset, our self-boosted method has rounded RMSE values for the lag times $3, 6, 12$ are $98, 86$, and $89$, respectively. In this metric setting, the proposed method outperforms the strongest baseline methods which is Seq2seq whose rounded RMSE value for lag time $3$ is $152$, equivalent to $36\%$ improvement; Dilated CNN whose rounded RMSE values for lag time $6$ is $124$, lag time $12$ is $112$, equivalent to $31\%$ and $21\%$ improvements, respectively. For the air temperature dataset, our self-boosted method also beats other deep learning models at time lag $3$ and $6$. The exchange rates dataset also shares the same result pattern where our method outperforms other methods on lag time $6$ and $12$. It is slightly behind the Dilated-CNN model at lag time $3$ while the two models still have R-squared values above $99\%$. In addition, we average RMSE across all lag times to generalize the performance comparison of all methods. Figure \ref{fig:avg-rmse-comparison} shows that our self-boosted model outperforms all the baselines. 
\begin{table}[htbp]
  \centering
  \caption{Comparison of the average RMSE performance of the proposed model and the average performance of the best models in each lag time. ABBM means Average Best Baseline Methods.}
    \begin{tabular}{l|c|c|c}
    \toprule
          & Electricity & Temperature & Exchange Rates \\
    \midrule
    ABBM  & 129.1706     &  1.7943     & 0.0095 \\
    \midrule
    \textbf{Self-boosted} & \textbf{91.1300}     & \textbf{1.5910}     & \textbf{0.0082} \\
    \bottomrule
    \end{tabular}%
  \label{tab:performanc-average}%
\end{table}%

Furthermore, we select the best baseline method on each lag time performance, then average them for each dataset (so called ABBM method in the resulting table). The result is then compared with our proposed model average performance across all lag times for those datasets. Table \ref{tab:performanc-average} displays the average results. The RMSEs of our self-boosted model are $91.1300, 1.5910$, and $0.0082$ while the RMSEs of the average best models are $129.1708, 1.7943$, and $0.0095$  on the datasets electricity consumption, air temperature, and exchange rates, respectively. This result confirms that our self-boosted model consistently outperforms those state-of-the-art models.

\subsection{The Role of Multi-task and Multi-view Learning}
To evaluate the combination of multi-task and multi-view learning approach, we created a multi-task learning model (named MTL) that has a similar architecture of the self-boosted model in which the multi-view component is ignored. In addition, we also created a multi-view learning model (named MTV) by dropping the auxiliary tasks from the proposed model. Then we measure the performance of these variants and present the results in Figure \ref{fig:comparison-mtl-mtv-all}. 

Figure \ref{fig:comparison-mtl-mtv-all} shows normalized RMSE values of the MTL. model, MTV. model and the self-boosted model across all datasets with respect to the selected time lags $\{1, 3, 6, 12\}$. From the figure, the self-boosed model outperforms the two variants on electricity consumption dataset. The MTV model has comparable RMSE at time lag $1$, but the results of other time lags still show the superiority for the proposed model.  In addition, on the air temperature and exchange rate datasets, our model performs three to four times better than the MTL variant. Overall, our self-boosted model outperforms its variants and has the lowest normalized RMSE values in each time lag across the three datasets.


\subsection{Understanding the Importance of Intrinsic Mode Functions}
To better understanding the importance of the Intrinsic Mode Functions in self-boosted mechanism, we conduct an evaluation of the forecasting performance by sorting its similarity with the original time series in descending order. Then, these IMFs are fed into the self-boosted model one at a time from the most similar IMF to the least one. To reflect the influence of the IMF, we calculate RMSE coefficients from the RMSEs measured from the model performance. RMSE coefficient $coeff_i$ for the IMF $i^{th}$ is computed as:
\begin{equation}
    coeff_i = \frac{RMSE_i}{\sum_{j=1}^{N} RMSE_j}
\end{equation}
where $N$ is the number of IMFs, $RMSE_i$ is the RMSE after including IMF $i^{th}$ into the model. The lower the value of the RMSE coefficient, the higher the importance of the intrinsic mode function in the model performance.

Figure \ref{fig:performance-imfs} presents the RMSE coefficient after adding each IMF one at a time tested on the three datasets. With electricity consumption and air temperature datasets, we see that adding more IMFs until the red point (the point in which an IMF is added for multi-view learning) helps reducing RMSE coefficient. It indicates that the model performance is improved via multi-task learning. Meanwhile, the exchange rates dataset show a slight negative performance on the performance after adding IMF $3^{th}$. From the red point onward, we noticed a decreasing trend of the RMSE coefficient on the electricity consumption dataset, but an increasing trend with the air temperature dataset, and a fluctuation on exchange rate dataset. These behaviors indicate that some IMFs have a negative effect while some other IMFs have a positive effect to the model performance. It entails that a better feature selection can be done to select more proper intrinsic mode functions for both multi-task learning and multi-view learning. We can drop the IMFs which cause an increment of the RMSE coefficient and keep the ones that lead to decrements of RMSE coefficient values. In other words, the overall performance of the model does improve for simple feature selection with k-mean clustering algorithm. However, we can still enhance the model performance with a better feature selection strategy.

\section{Conclusion}
\label{sec:conclusion}
In this paper, we presented a novel self-boosted deep learning model for time series forecasting. The proposed model co-trains multi-task learning and multi-view learning to enhance the forecasting performance. The learning features come from intrinsic mode functions which are generated by the ensemble empirical mode decomposition (EEMD) method in the signal processing domain. The multi-task learning part learns from related intrinsic mode functions while the multi-view learning component learns from less related ones. Three public datasets: electricity consumption, air temperature and exchange rates are used to evaluate the forecasting results. The experimental results demonstrate that our proposed self-boosted model outperforms several state-of-the-art baseline methods on all the datasets. Future work will continue to explore the intrinsic mode functions selection strategy, so that the negative impact on the model performance will be removed, and the network computation can be more efficient.

\bibliographystyle{aaai}
\bibliography{reference}

\end{document}